# Substructure Discovery Using Minimum Description Length and Background Knowledge

**Diane J. Cook**                                             COOK@CSE.UTA.EDU
**Lawrence B. Holder**                                        HOLDER@CSE.UTA.EDU
*Department of Computer Science Engineering*
*Box 19015*
*University of Texas at Arlington*
*Arlington, TX 76019 USA*

## Abstract

The ability to identify interesting and repetitive substructures is an essential component to discovering knowledge in structural data. We describe a new version of our SUB-DUE substructure discovery system based on the minimum description length principle. The SUBDUE system discovers substructures that compress the original data and represent structural concepts in the data. By replacing previously-discovered substructures in the data, multiple passes of SUBDUE produce a hierarchical description of the structural regularities in the data. SUBDUE uses a computationally-bounded inexact graph match that identifies similar, but not identical, instances of a substructure and finds an approximate measure of closeness of two substructures when under computational constraints. In addition to the minimum description length principle, other background knowledge can be used by SUBDUE to guide the search towards more appropriate substructures. Experiments in a variety of domains demonstrate SUBDUE's ability to find substructures capable of compressing the original data and to discover structural concepts important to the domain.

## 1. Introduction

The large amount of data collected today is quickly overwhelming researchers' abilities to interpret the data and discover interesting patterns within the data. In response to this problem, a number of researchers have developed techniques for discovering concepts in databases. These techniques work well for data expressed in a non-structural, attribute-value representation, and address issues of data relevance, missing data, noise and uncertainty, and utilization of domain knowledge. However, recent data acquisition projects are collecting structural data describing the relationships among the data objects. Correspondingly, there exists a need for techniques to analyze and discover concepts in structural databases.

One method for discovering knowledge in structural data is the identification of common substructures within the data. The motivation for this process is to find substructures capable of compressing the data and to identify conceptually interesting substructures that enhance the interpretation of the data. Substructure discovery is the process of identifying concepts describing interesting and repetitive substructures within structural data. Once discovered, the substructure concept can be used to simplify the data by replacing instances of the substructure with a pointer to the newly discovered concept. The discovered substructure concepts allow abstraction over detailed structure in the original data and provide new, relevant attributes for interpreting the data. Iteration of the substructure discovery and replacement process constructs a hierarchical description of the structural data in terms of the discovered substructures. This hierarchy provides varying levels of interpretation that can be accessed based on the goals of the data analysis.

We describe a system called Subdue (Holder, Cook, & Bunke, 1992; Holder & Cook, 1993) that discovers interesting substructures in structural data based on the minimum description length principle. The Subdue system discovers substructures that compress the original data and represent structural concepts in the data. By replacing previously-discovered substructures in the data, multiple passes of Subdue produce a hierarchical description of the structural regularities in the data. Subdue uses a computationally-bounded inexact graph match that identifies similar, but not identical, instances of a substructure and finds an approximate measure of closeness of two substructures when under computational constraints. In addition to the minimum description length principle, other background knowledge can be used by Subdue to guide the search towards more appropriate substructures.

The following sections describe the approach in detail. Section 2 describes the process of substructure discovery and introduces needed definitions. Section 3 compares the Subdue discovery system to other work found in the literature. Section 4 introduces the minimum description length encoding used by this approach, and Section 5 presents the inexact graph match algorithm employed by Subdue. Section 6 describes methods of incorporating background knowledge into the substructure discovery process. The experiments detailed in Section 7 demonstrate Subdue's ability to find substructures that compress the data and to re-discover known concepts in a variety of domains. Section 8 details the hierarchical discovery process. We conclude with observations and directions for future research.

## 2. Substructure Discovery

The substructure discovery system represents structured data as a labeled graph. Objects in the data map to vertices or small subgraphs in the graph, and relationships between objects map to directed or undirected edges in the graph. A *substructure* is a connected subgraph within the graphical representation. This graphical representation serves as input to the substructure discovery system. Figure 1 shows a geometric example of such an input graph. The objects in the figure (e.g., `T1`, `S1`, `R1`) become labeled vertices in the graph, and the relationships (e.g., `on(T1,S1)`, `shape(C1,circle)`) become labeled edges in the graph. The graphical representation of the substructure discovered by Subdue from this data is also shown in Figure 1.

An *instance* of a substructure in an input graph is a set of vertices and edges from the input graph that match, graph theoretically, to the graphical representation of the substructure. For example, the instances of the substructure in Figure 1 are shown in Figure 2.

The substructure discovery algorithm used by Subdue is a computationally-constrained beam search. The algorithm begins with the substructure matching a single vertex in the graph. Each iteration through the algorithm selects the best substructure and expands the instances of the substructure by one neighboring edge in all possible ways. The new unique generated substructures become candidates for further expansion. The algorithm searches for the best substructure until all possible substructures have been considered or the total amount of computation exceeds a given limit. The evaluation of each substructure is guided by the MDL principle and other background knowledge provided by the user.

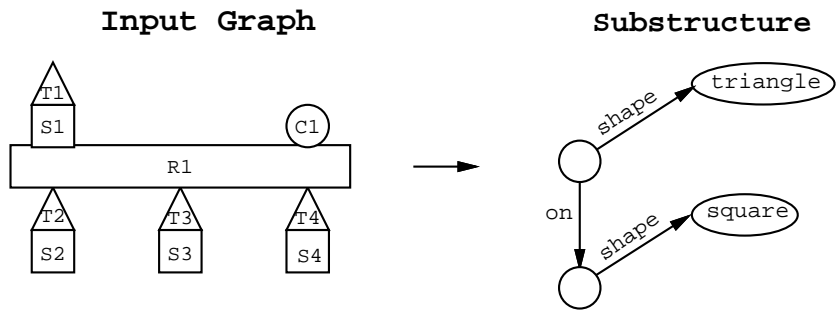

Figure 1: Example substructure in graph form.

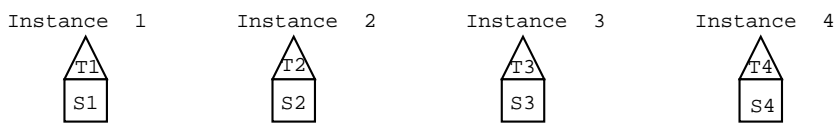

Figure 2: Instances of the substructure.

Typically, once the description length of an expanding substructure begins to increase, further expansion of the substructure will not yield a smaller description length. As a result, SUBDUE makes use of an optional pruning mechanism that eliminates substructure expansions from consideration when the description lengths for these expansions increases.

## 3. Related Work

Several approaches to substructure discovery have been developed. Winston's ARCH program (Winston, 1975) discovers substructures in order to deepen the hierarchical description of a scene and to group objects into more general concepts. The ARCH program searches for two types of substructure in the blocks-world domain. The first type involves a sequence of objects connected by a chain of similar relations. The second type involves a set of objects each having a similar relationship to some "grouping" object. The main difference between the substructure discovery procedures used by the ARCH program and SUBDUE is that the ARCH program is designed specifically for the blocks-world domain. For instance, the sequence discovery method looks for `supported-by` and `in-front-of` relations only. SUBDUE's substructure discovery method is domain independent, although the inclusion of domain-specific knowledge would improve SUBDUE's performance.

Motivated by the need to construct a knowledge base of chemical structures, Levinson (Levinson, 1984) developed a system for storing labeled graphs in which individual graphs are represented by the set of vertices in a universal graph. In addition, the individual graphs are maintained in a partial ordering defined by the `subgraph-of` relation, which improves the performance of graph comparisons. The universal graph representation provides a method for compressing the set of graphs stored in the knowledge base. Subgraphs of the universal graph used by several individual graphs suggest common substructure in the individual graphs. One difference between the two approaches is that Levinson's system is designed to incrementally process smaller individual graphs; whereas, SUBDUE processes larger graphs all at once. Also, Levinson's system discovers common substructure only as an indirect result of the universal graph construction; whereas, SUBDUE's main goal is to discover and output substructure definitions that reduce the minimum description length encoding of the graph. Finally, the `subgraph-of` partial ordering used by Levinson's system is not included in SUBDUE, but maintaining this partial ordering would improve the performance of the graph matching procedure by pruning the number of possible matching graphs.

Segen (Segen, 1990) describes a system for storing graphs using a probabilistic graph model to represent subsets of the graph. Alternative models are evaluated based on a minimum description length measure of the information needed to represent the stored graphs using the model. In addition, Segen's system clusters the graphs into classes based on minimizing the description length of the graphs according to the entire clustering. Apart from the probabilistic representation, Segen's approach is similar to Levinson's system in that both methods take advantage of commonalities in the graphs to assist in graph storage and matching. The probabilistic graphs contain information for identifying common substructure in the exact graphs they represent. The portion of the probabilistic graph with high probability defines a substructure that appears frequently in the exact graphs. This notion was not emphasized in Segen's work, but provides an alternative method to substructure discovery by clustering subgraphs of the original input graphs. As with Levinson's approach, graphs are processed incrementally, and substructure is found across several graphs, not within a single graph as in SUBDUE.

The LABYRINTH system (Thompson & Langley, 1991) extends the COBWEB incremental conceptual clustering system (Fisher, 1987) to handle structured objects. LABYRINTH uses COBWEB to form hierarchical concepts of the individual objects in the domain based on their primitive attributes. Concepts of structured objects are formed in a similar manner using the individual objects as attributes. The resulting hierarchy represents a componential model of the structured objects. Because COBWEB's concepts are probabilistic, LABYRINTH produces probabilistic models of the structured objects, but with an added hierarchical organization. The upper-level components of the structured-object hierarchy produced by LABYRINTH represent substructures common to the examples. Therefore, although not the primary focus, LABYRINTH is discovering substructure, but in a more constrained context than the general graph representation used by SUBDUE.

Conklin et al. (Conklin & Glasgow, 1992) have developed the I-MEM system for constructing an image hierarchy, similar to that of LABYRINTH, used for discovering common substructures in a set of images and for efficient retrieval of images similar to a given image. Images are expressed in terms of a set of relations defined by the user. Specific and general (conceptual) images are stored in the hierarchy based on a subsumption relation similar to Levinson's **subgraph-of** partial ordering. Image matching utilizes a transformational approach (similar to Subdue's inexact graph match) as a measure of image closeness.

As with the approaches of Segen and Levinson, i-mem is designed to process individual images. Therefore, the general image concepts that appear higher in i-mem's hierarchy will represent common substructures across several images. Subdue is designed to discover common substructures within a single image. Subdue can mimic the individual approach of these systems by processing a set of individual images as one disconnected graph. The substructures found will be common to the individual images. The hierarchy also represents a componential view of the images. This same view can be constructed by Subdue using multiple passes over the graph after replacing portions of the input graph with substructures discovered during previous passes. i-mem has performed well in a simple chess domain and molecular chemistry domains (Conklin & Glasgow, 1992). However, i-mem requires domain-specific relations for expressing images in order for the hierarchy to find relevant substructures and for image matching to be efficient. Again, maintaining the concepts (images, graphs) in a partially-ordered hierarchy improves the efficiency of matching and retrieval, and suggests a possible improvement to Subdue.

The CLiP system (Yoshida, Motoda, & Indurkhya, 1993) for graph-based induction is more similar to Subdue than the previous systems. CLiP iteratively discovers patterns in graphs by expanding and combining patterns discovered in previous iterations. Patterns are grouped into views based on their collective ability to compress the original input graph. During each iteration CLiP uses existing views to contract the input graph and then considers adding to the views new patterns consisting of two vertices and an edge from the contracted graph. The compression of the new proposed views is estimated, and the best views (according to a given beam width) are retained for the next iteration.

CLiP discovers substructures (patterns) differently than Subdue. First, CLiP produces a set of substructures that collectively compress the input graph; whereas, Subdue produces only single substructures evaluated using the more principled minimum description length. CLiP has the ability to grow substructures agglomeratively (i.e., merging two substructures together); whereas, Subdue always produces new substructures using incremental growth along one new edge. CLiP initially estimates the compression value of new views based on the compression value of the parent view; whereas, Subdue performs an expensive exact measurement of compression for each new substructure. Finally, CLiP employs an efficient graph match based on graph identity, not graph isomorphism as in Subdue. Graph identity assumes an ordering over the incident edges of a vertex and does not consider all possible mappings when looking for occurrences of a pattern in an input graph. These differences in CLiP suggest possible enhancements to Subdue.

Research in pattern recognition has begun to investigate the use of graphs and graph grammars as an underlying representation for structural problems (Schalkoff, 1992). Many results in grammatical inference are applicable to constrained classes of graphs (e.g., trees) (Fu, 1982; Miclet, 1986). The approach begins with a set of sample graphs and produces a generalized graph grammar capable of deriving the original sample graphs and many others. The production rules of this general grammar capture regularities (substructures) in the sample graphs. Jeltsch and Kreowski (Jeltsch & Kreowski, 1991) describe an approach that begins with a maximally-specific grammar and iteratively identifies common subgraphs in the right-hand sides of the production rules. These common subgraphs are used to form new, more general production rules. Although their method does not address the underlying combinatorial nondeterminism, heuristic approaches could provide a feasible method for extracting substructures in the form of graph grammars. Furthermore, the graph grammar production-rule may provide a suitable representation for background knowledge during the substructure discovery process.

## 4. Minimum Description Length Encoding of Graphs

The minimum description length principle (MDLP) introduced by Rissanen (Rissanen, 1989) states that the best theory to describe a set of data is that theory which minimizes the description length of the entire data set. The MDL principle has been used for decision tree induction (Quinlan & Rivest, 1989), image processing (Pednault, 1989; Pentland, 1989; Leclerc, 1989), concept learning from relational data (Derthick, 1991), and learning models of non-homogeneous engineering domains (Rao & Lu, 1992).

We demonstrate how the minimum description length principle can be used to discover substructures in complex data. In particular, a substructure is evaluated based on how well it can compress the entire dataset using the minimum description length. We define the minimum description length of a graph to be the number of bits necessary to completely describe the graph.

According to the minimum description length (MDL) principle, the theory that best accounts for a collection of data is the one that minimizes $I(S) + I(G|S)$, where $S$ is the discovered substructure, $G$ is the input graph, $I(S)$ is the number of bits required to encode the discovered substructure, and $I(G|S)$ is the number of bits required to encode the input graph $G$ with respect to $S$.

The graph connectivity can be represented by an adjacency matrix. Consider a graph that has $n$ vertices, which are numbered $0, 1, \ldots, n - 1$. An $n \times n$ adjacency matrix $A$ can be formed with entry $A[i, j]$ set to 0 or 1. If $A[i, j] = 0$, then there is no connection from vertex $i$ to vertex $j$. If $A[i, j] = 1$, then there is at least one connection from vertex $i$ to vertex $j$. Undirected edges are recorded in only one entry of the matrix. The adjacency matrix for the graph in Figure 3 is shown below.

$$
\begin{array}{r}
x \\
triangle \\
y \\
square \\
r \\
rectangle
\end{array}
\left[
\begin{array}{cccccc}
0 & 1 & 1 & 0 & 0 & 0 \\
0 & 0 & 0 & 0 & 0 & 0 \\
0 & 0 & 0 & 1 & 1 & 0 \\
0 & 0 & 0 & 0 & 0 & 0 \\
0 & 0 & 0 & 0 & 0 & 1 \\
0 & 0 & 0 & 0 & 0 & 0
\end{array}
\right]
$$

The encoding of the graph consists of the following steps. We assume that the decoder has a table of the $l_u$ unique labels in the original graph $G$.

1. Determine the number of bits *vbits* needed to encode the vertex labels of the graph. First, we need $(\lg v)$ bits to encode the number of vertices $v$ in the graph. Then, encoding the labels of all $v$ vertices requires $(v \lg l_u)$ bits. We assume the vertices are specified in the same order they appear in the adjacency matrix. The total number of bits to encode the vertex labels is

$$vbits = \lg v + v \lg l_u$$

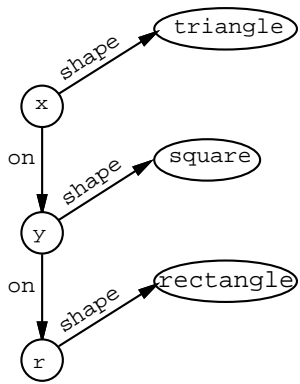

Figure 3: MDL example graph.

For the example in Figure 3, $v = 6$, and we assume that there are $l_u = 8$ unique labels in the original graph. The number of bits needed to encode these vertices is $\lg 6 + 6 \lg 8 = 20.58$ bits.

2. Determine the number of bits *rbits* needed to encode the rows of the adjacency matrix $A$. Typically, in large graphs, a single vertex has edges to only a small percentage of the vertices in the entire graph. Therefore, a typical row in the adjacency matrix will have much fewer than $v$ 1s, where $v$ is the total number of vertices in the graph. We apply a variant of the coding scheme used by (Quinlan & Rivest, 1989) to encode bit strings with length $n$ consisting of $k$ 1s and $(n - k)$ 0s, where $k \ll (n - k)$. In our case, row $i$ ($1 \leq i \leq v$) can be represented as a bit string of length $v$ containing $k_i$ 1s. If we let $b = \max_i k_i$, then the $i^{th}$ row of the adjacency matrix can be encoded as follows.

(a) Encoding the value of $k_i$ requires $\lg(b + 1)$ bits.

(b) Given that only $k_i$ 1s occur in the row bit string of length $v$, only $\binom{v}{k_i}$ strings of 0s and 1s are possible. Since all of these strings have equal probability of occurrence, $\lg \binom{v}{k_i}$ bits are needed to encode the positions of 1s in row $i$. The value of $v$ is known from the vertex encoding.

Finally, we need an additional $\lg(b + 1)$ bits to encode the number of bits needed to specify the value of $k_i$ for each row. The total encoding length in bits for the adjacency matrix is

$$
\begin{aligned}
rbits &= \lg(b + 1) + \sum_{i=1}^{v} \lg(b + 1) + \lg \binom{v}{k_i} \\
&= (v + 1)\lg(b + 1) \sum_{i=1}^{v} \lg \binom{v}{k_i}
\end{aligned}
$$

For the example in Figure 3, $b = 2$, and the number of bits needed to encode the adjacency matrix is $(7 \lg 3) + \lg \binom{9}{2} + \lg \binom{6}{6} + \lg \binom{9}{2} + \lg \binom{6}{6} + \lg \binom{9}{4} + \lg \binom{6}{6} = 21.49$ bits.

3. Determine the number of bits $ebits$ needed to encode the edges represented by the entries $A[i, j] = 1$ of the adjacency matrix $A$. The number of bits needed to encode entry $A[i, j]$ is $(\lg m) + e(i, j)[1 + \lg l_u]$, where $e(i, j)$ is the actual number of edges between vertex $i$ and $j$ in the graph and $m = \max_{i,j} e(i, j)$. The $(\lg m)$ bits are needed to encode the number of edges between vertex $i$ and $j$, and $[1 + \lg l_u]$ bits are needed per edge to encode the edge label and whether the edge is directed or undirected. In addition to encoding the edges, we need to encode the number of bits $(\lg m)$ needed to specify the number of edges per entry. The total encoding of the edges is

$$
\begin{aligned}
ebits &= \lg m + \sum_{i=1}^{v} \sum_{j=1}^{v} \lg m + e(i, j)[1 + \lg l_u] \\
&= \lg m + e(1 + \lg l_u) + \sum_{i=1}^{v} \sum_{j=1}^{v} A[i, j] \lg m \\
&= e(1 + \lg l_u) + (K + 1) \lg m
\end{aligned}
$$

where $e$ is the number of edges in the graph, and $K$ is the number of 1s in the adjacency matrix $A$. For the example in Figure 3, $e = 5$, $K = 5$, $m = 1$, $l_u = 8$, and the number of bits needed to encode the edges is $5(1 + \lg 8) + 6 \lg 1 = 20$.

The total encoding of the graph takes $(vbits + rbits + ebits)$ bits. For the example in Figure 3, this value is 62.07 bits.

Both the input graph and discovered substructure can be encoded using the above scheme. After a substructure is discovered, each instance of the substructure in the input graph is replaced by a single vertex representing the entire substructure. The discovered substructure is represented in $I(S)$ bits, and the graph after the substructure replacement is represented in $I(G|S)$ bits. Subdue searches for the substructure $S$ in graph $G$ minimizing $I(S) + I(G|S)$.

## 5. Inexact Graph Match

Although exact structure match can be used to find many interesting substructures, many of the most interesting substructures show up in a slightly different form throughout the data. These differences may be due to noise and distortion, or may just illustrate slight differences between instances of the same general class of structures. Consider the image shown in Figure 9. The pencil and the cube would make ideal substructures in the picture, but an exact match algorithm may not consider these as strong substructures, because they rarely occur in the same form and level of detail throughout the picture.

Given an input graph and a set of defined substructures, we want to find those subgraphs of the input graph that most closely resemble the given substructures. Furthermore, we want to associate a distance measure between a pair of graphs consisting of a given substructure and a subgraph of the input graph. We adopt the approach to inexact graph match given by Bunke and Allermann (Bunke & Allermann, 1983).

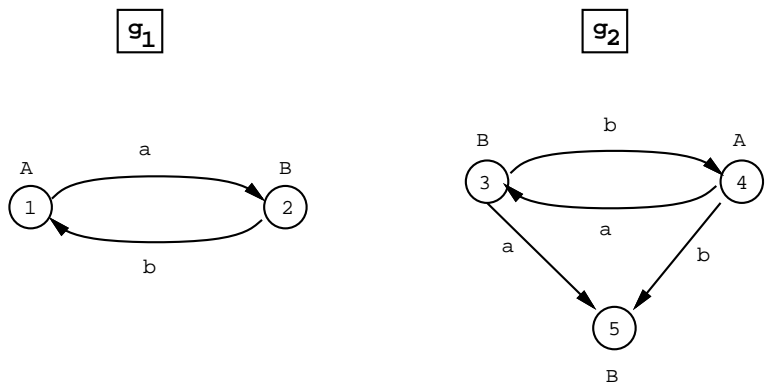

Figure 4: Two similar graphs $g_1$ and $g_2$.

In this inexact match approach, each distortion of a graph is assigned a cost. A distortion is described in terms of basic transformations such as deletion, insertion, and substitution of vertices and edges. The distortion costs can be determined by the user to bias the match for or against particular types of distortions.

An inexact graph match between two graphs $g_1$ and $g_2$ maps $g_1$ to $g_2$ such that $g_2$ is interpreted as a distorted version of $g_1$. Formally, an inexact graph match is a mapping $f : N_1 \rightarrow N_2 \cup \{\lambda\}$, where $N_1$ and $N_2$ are the sets of vertices of $g_1$ and $g_2$, respectively. A vertex $v \in N_1$ that is mapped to $\lambda$ (i.e., $f(v) = \lambda$) is deleted. That is, it has no corresponding vertex in $g_2$. Given a set of particular distortion costs as discussed above, we define the cost of an inexact graph match $cost(f)$, as the sum of the cost of the individual transformations resulting from $f$, and we define $matchcost(g_1, g_2)$ as the value of the least-cost function that maps graph $g_1$ onto graph $g_2$.

Given $g_1$, $g_2$, and a set of distortion costs, the actual computation of $matchcost(g_1, g_2)$ can be determined using a tree search procedure. A state in the search tree corresponds to a partial match that maps a subset of the vertices of $g_1$ to a subset of the vertices in $g_2$. Initially, we start with an empty mapping at the root of the search tree. Expanding a state corresponds to adding a pair of vertices, one from $g_1$ and one from $g_2$, to the partial mapping constructed so far. A final state in the search tree is a match that maps all vertices of $g_1$ to $g_2$ or to $\lambda$. The complete search tree of the example in Figure 4 is shown in Figure 5. For this example we assign a value of 1 to each distortion cost. The numbers in circles in this figure represent the cost of a state. As we are eventually interested in the mapping with minimum cost, each state in the search tree gets assigned the cost of the partial mapping that it represents. Thus the goal state to be found by our tree search procedure is the final state with minimum cost among all final states. From Figure 5 we conclude that the minimum cost inexact graph match of $g_1$ and $g_2$ is given by the mapping $f(1) = 4$, $f(2) = 3$. The cost of this mapping is 4.

Given graphs $g_1$ with $n$ vertices and $g_2$ with $m$ vertices, $m \geq n$, the complexity of the full inexact graph match is $O(n^{m+1})$. Because this routine is used heavily throughout the discovery and evaluation process, the complexity of the algorithm can significantly degrade the performance of the system.

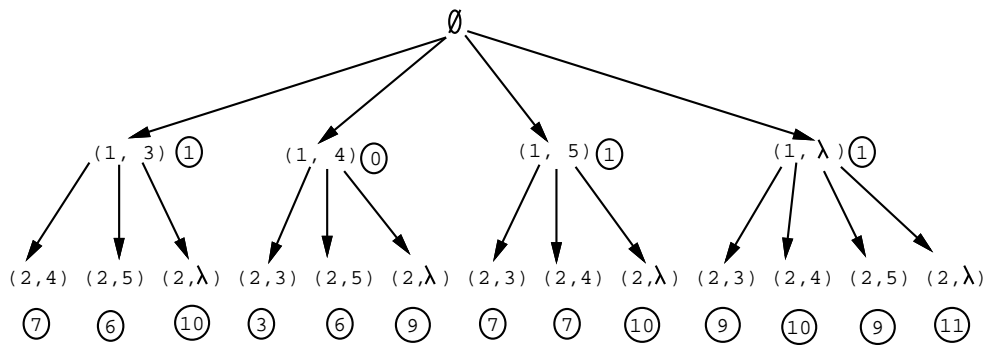

Figure 5: Search tree for computing matchcost($g_1$,$g_2$) from Figure 4.

To improve the performance of the inexact graph match algorithm, we extend Bunke's approach by applying a branch-and-bound search to the tree. The cost from the root of the tree to a given node is computed as described above. Nodes are considered for pairings in order from the most heavily connected vertex to the least connected, as this constrains the remaining match. Because branch-and-bound search guarantees an optimal solution, the search ends as soon as the first complete mapping is found.

In addition, the user can place a limit on the number of search nodes considered by the branch-and-bound procedure (defined as a function of the size of the input graphs). Once the number of nodes expanded in the search tree reaches the defined limit, the search resorts to hill climbing using the cost of the mapping so far as the measure for choosing the best node at a given level. By defining such a limit, significant speedup can be realized at the expense of accuracy for the computed match cost.

Another approach to inexact graph match would be to encode the difference between two graphs using the MDL principle. Smaller encodings would indicate a lower match cost between the two graphs. We leave this as a future research direction.

## 6. Guiding the Discovery Process with Background Knowledge

Although the principle of minimum description length is useful for discovering substructures that maximize compression of the data, scientists may realize more benefit from the discovery of substructures that exhibit other domain-specific and domain-independent characteristics.

To make SUBDUE more powerful across a wide variety of domains, we have added the ability to guide the discovery process with background knowledge. Although the minimum description length principle still drives the discovery process, the background knowledge can be used to input a bias toward certain types of substructures. This background knowledge is encoded in the form of rules for evaluating substructures, and can represent domain-independent or domain-dependent rules. Each time a substructure is evaluated, these input rules are used to determine the value of the substructure under consideration. Because only the most-favored substructures are kept and expanded, these rules bias the discovery process of the system.

Each background rule can be assigned a positive, zero, or negative weight, that biases the procedure toward a type of substructure, eliminates the use of the rule, or biases the procedure away from a type of substructure, respectively. The value of a substructure is defined as the description length (DL) of the input graph using the substructure multiplied by the weighted value of each background rule from a set of rules $R$ applied to the substructure.

$$value(s) = DL(G, s) \times \prod_{r=1}^{|R|} \text{rule}_r(s)^{e_r} \tag{1}$$

Three domain-independent heuristics that have been incorporated as rules into the SUB-DUE system are compactness, connectivity, and coverage. For the definitions of these rules, we will let $G$ represent the input graph, $s$ represent a substructure in the graph, and $I$ represent the set of instances of the substructure $s$ in $G$. The instance weight $w$ of an instance $i \in I$ of a substructure $s$ is defined to be

$$w(i, s) = 1 - \frac{matchcost(i, s)}{size(i)}, \tag{2}$$

where $size(i) = \#vertices(i) + \#edges(i)$. If the match cost is greater than the size of the larger graph, then $w(i, s) = 0$. The instance weights are used in these rules to compute a weighted average over instances of a substructure. A value of 1 is added to each formula so that the exponential weights can be used to control the rule's significance.

The first rule, *compactness*, is a generalization of Wertheimer's *Factor of Closure*, which states that human attention is drawn to closed structures (Wertheimer, 1939). A closed substructure has at least as many edges as vertices, whereas a non-closed substructure has fewer edges than vertices (Prather, 1976). Thus, closed substructures have a higher compactness value. Compactness is defined as the weighted average of the ratio of the number of edges in the substructure to the number of vertices in the substructure.

$$compactness(s) = 1 + \frac{1}{|I|} \sum_{i \in I} w(i, s) \times \frac{\#edges(i)}{\#vertices(i)} \tag{3}$$

The second rule, *connectivity*, measures the amount of external connection in the instances of the substructure. The connectivity rule is a variant of Wertheimer's *Factor of Proximity* (Wertheimer, 1939), and is related to earlier numerical clustering techniques (Zahn, 1971). These works demonstrate the human preference for "isolated" substructures, that is, substructures that are minimally related to adjoining structure. Connectivity measures the "isolation" of a substructure by computing the inverse of the average number of external connections over all the weighted instances of the substructure in the input graph. An *external connection* is defined here as an edge that connects a vertex in the substructure to a vertex outside the substructure. The formula for determining the connectivity of a substructure $s$ with instances $I$ in the input graph $G$ is given below.

$$connectivity(s) = 1 + \left[ \frac{1}{|I|} \sum_{i \in I} w(i,s) \times num\_external\_conns(i) \right]^{-1} \quad (4)$$

The third rule, *coverage*, measures the fraction of structure in the input graph described by the substructure. The coverage rule is motivated from research in inductive learning and provides that concept descriptions describing more input examples are considered better (Michalski & Stepp, 1983). Although MDL measures the amount of structure, the coverage rule includes the relevance of this savings with respect to the size of the entire input graph. Coverage is defined as the number of unique vertices and edges in the instances of the substructure divided by the total number of vertices and edges in the input graph. In this formula, the *unique_structure(i)* of an instance $i$ is the number of vertices and edges in $i$ that have not already appeared in previous instances in the summation.

$$coverage(s) = 1 + \frac{\sum_{i \in I} w(i,s) \times unique\_structure(i)}{size(G)} \quad (5)$$

Domain-dependent rules can also be used to guide the discovery process in a domain where scientists can contribute their expertise. For example, CAD circuits generally consist of two types of components, active and passive components. The active components are the main driving components. Identifying the active components is the first step in understanding the main function of the circuit. To add this knowledge to Subdue we include a rule that assigns higher values to substructures (circuit components) representing active components and lower values to substructures representing passive components. Since the active components have higher scores, they are expected to be selected. The system can then focus the attention on the active components which will be expanded to the functional substructures.

Another method of biasing the discovery process with background knowledge is to let background rules affect the prior probabilities of possible substructures. However, choosing the appropriate prior probabilities to express desired properties of substructures is difficult, but indicates a future direction for the inclusion of background knowledge into the substructure discovery process.

## 7. Experiments

The experiments in this section evaluate Subdue's substructure discovery capability in several domains, including chemical compound analysis, scene analysis, CAD circuit design analysis, and analysis of an artificially-generated structural database.

Two goals of our substructure discovery system are to find substructures that can reduce the amount of information needed to describe the data, and to find substructures that are considered interesting for the given database. As a result, we evaluate the Subdue system in this section along these two criteria. First, we measure the amount of compression that Subdue provides across a variety of databases. Second, we use the Subdue system with the additional background knowledge rules to re-discover substructures that have been identified as interesting by experts in each specific domain. Section 7.1 describes the domains used in these experiments, and Section 7.2 presents the experimental results.

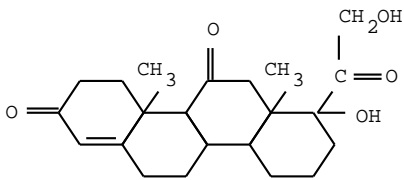

Figure 6: Cortisone.

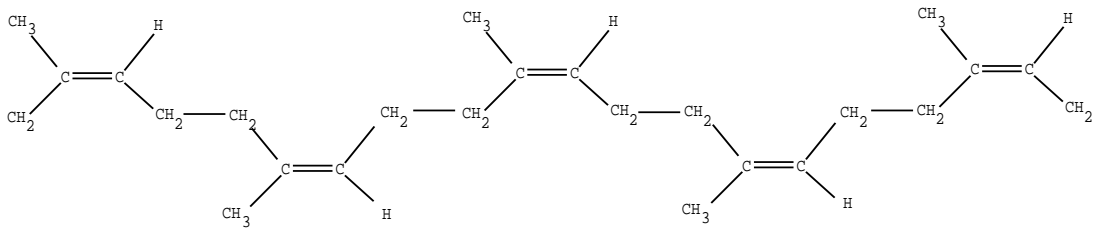

Figure 7: Natural rubber (all-cis polyisoprene).

## 7.1 Domains

### 7.1.1 CHEMICAL COMPOUND ANALYSIS

Chemical compounds are rich in structure. Identification of the common and interesting substructures can benefit scientists by identifying recurring components, simplying the data description, and focusing on substructures that stand out and merit additional attention.

Chemical compounds are represented graphically by mapping individual atoms, such as carbon and oxygen, to labeled vertices in the graph, and by mapping bonds between the atoms onto labeled edges in the graph. Figures 6, 7, and 8 show the graphs representing the chemical compound databases for cortisone, rubber, and a portion of a DNA molecule.

### 7.1.2 SCENE ANALYSIS

Images and scene descriptions provide a rich source of structure. Images that humans encounter, both natural and synthesized, have many structured subcomponents that draw our attention and that help us to interpret the data or the scene.

Discovering common structures in scenes can be useful to a computer vision system. First, automatic substructure discovery can help a system interpret an image. Instead of working from low-level vertices and edges, SUBDUE can provide more abstract structured components, resulting in a hierarchical view of the image that the machine can analyze at many levels of detail and focus, depending on the goal of the analysis. Second, substructure discovery that makes use of an inexact graph match can help identify objects in a 2D image of a 3D scene where noise and orientation differences are likely to exist. If an object appears often in the scene, the inexact graph match driving the SUBDUE system may capture slightly different views of the same object. Although an object may be difficult to identify from just one 2D picture, Subdue will match instances of similar objects, and the differences between these instances can provide additional information for identification. Third, substructure discovery can be used to compress the image. Replacing common interesting substructures by a single vertex simplifies the image description and reduces the amount of storage necessary to represent the image.

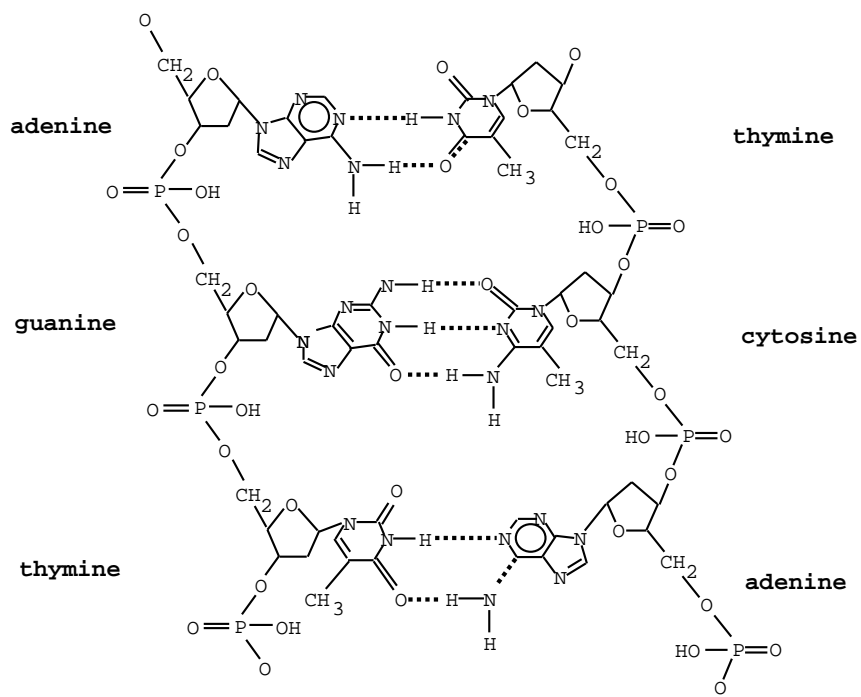

Figure 8: Portion of a DNA molecule.

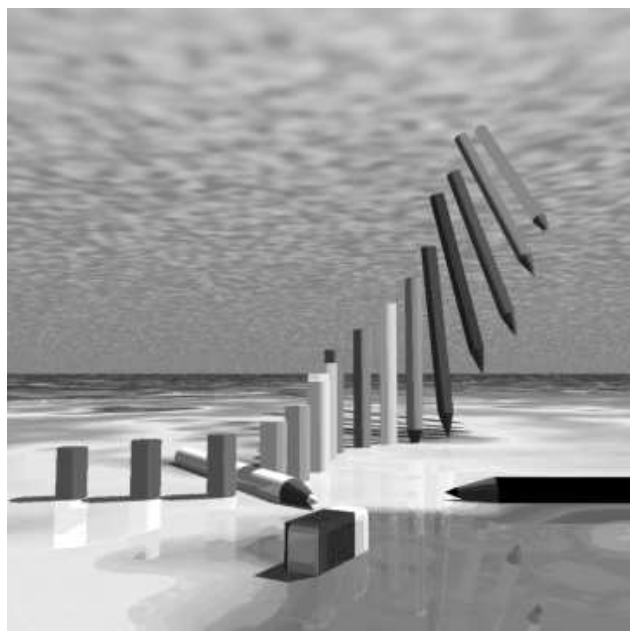

Figure 9: Scene analysis example.

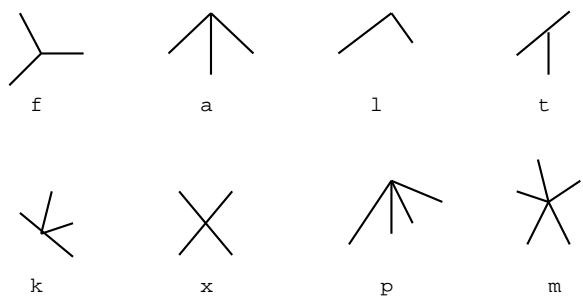

Figure 10: Possible vertices and labels.

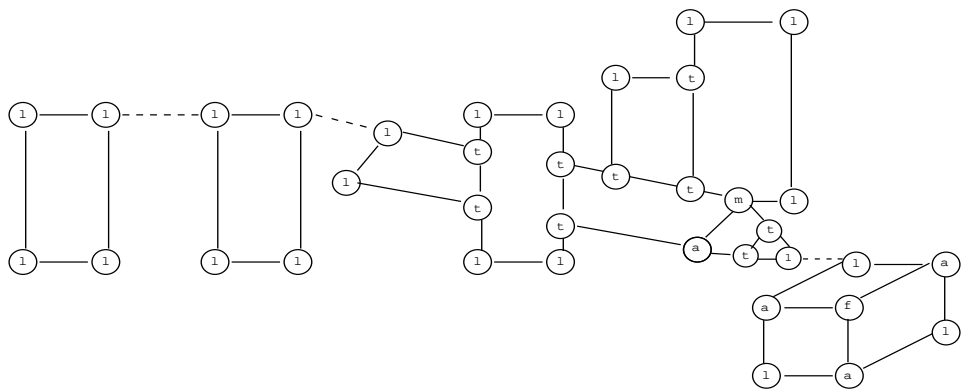

Figure 11: Portion of graph representing scene in Figure 4.

To apply Subdue to image data, we extract edge information from the image and construct a graph representing the scene. The graph representation consists of eight types of vertices and two types of arcs (*edge* and *space*). The vertex labels (*f*, *a*, *l*, *t*, *k*, *x*, *p*, and *m*) follow the Waltz labelings (Waltz, 1975) of junctions of edges in the image and represent the types of vertices shown in Figure 10. An *edge* arc represents the edge of an object in the image, and a *space* arc links non-connecting objects together. The *edge* arcs represent an edge in the scene that connects two vertices, and the *space* arcs connect the closest vertices from two disjoint neighboring objects. Distance, curve, and angle information has not been included in the graph representation, but can be added to give additional information about the scene. Figure 11 shows the graph representation of a portion of the scene depicted in Figure 9. In this figure, the *edge* arcs are solid and the *space* arcs are dashed.

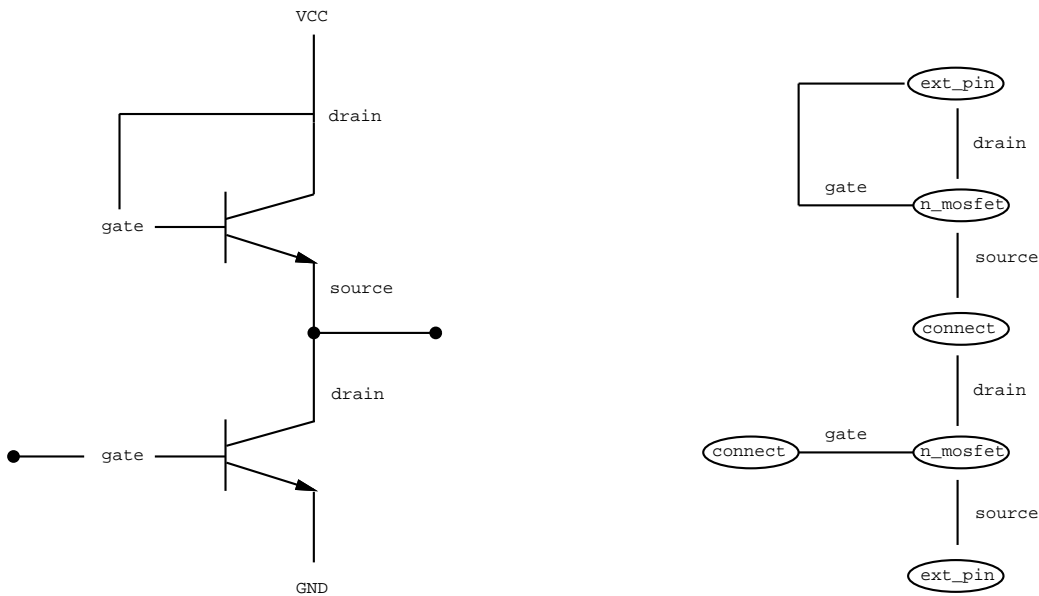

Figure 12: Amplifier circuit and graph representation.

### 7.1.3 CAD CIRCUIT ANALYSIS

In this domain, we employ SUBDUE to find circuit components in CAD circuit data. Discovery of substructures in circuit data can be a valuable tool to an engineer who is attempting to identify common reusable parts in a circuit layout. Replacing individual components in the circuit description by larger substructure descriptions will also simplify the representation of the circuit.

The data for the circuit domain was obtained from National Semiconductor, and consists of a set of components making up a circuit as output by the Cadence Design System. The particular circuit used for this experiment is a portion of an analog-to-digital converter. Figure 12 presents a circuit for an amplifier and gives the corresponding graph representation.

### 7.1.4 ARTIFICIAL DOMAIN

In the final domain, we artificially generate graphs to evaluate SUBDUE's ability to discover substructures capable of compressing the graph. Four substructures are created of varying sizes with randomly-selected vertices and edges (see Figure 13). The name of a substructure reflects the number of vertices and edges in its graph representation. Next, these substructures are embedded in larger graphs whose size is 15 times the size of the substructure. The graphs vary across four parameters: number of possible vertex and edge labels (one times and two times the number of labels used in the substructure), connectivity of the substructure (1 or 2 external connections), coverage of the instances (60% and 80%), and the amount of distortion in the instances (0, 1 or 2 distortions). This yields a total of 96 graphs (24 for each different substructure).

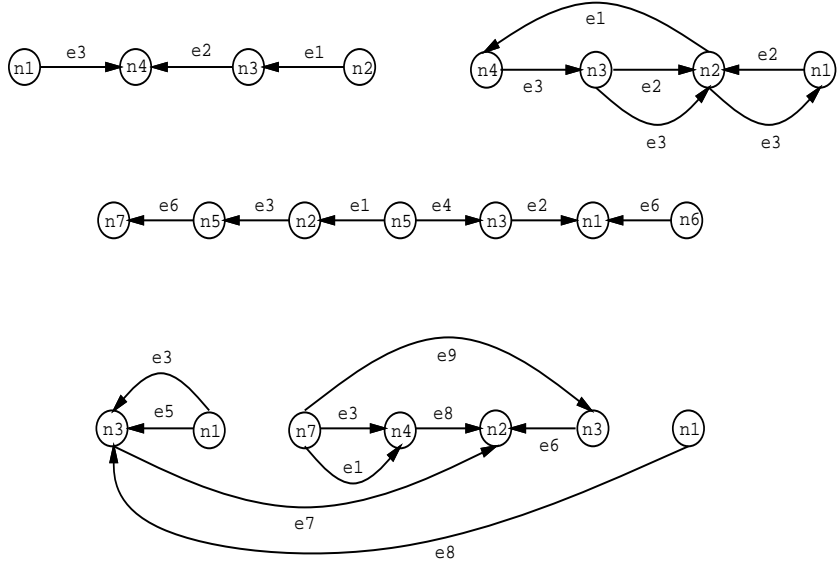

Figure 13: Four artificial substructures used to evaluate SUBDUE.

## 7.2 Experimental Results

### 7.2.1 EXPERIMENT 1: DATA COMPRESSION

In the first experiment, we test SUBDUE's ability to compress a structural database. Using a beam width of 4 and SUBDUE's pruning mechanism, we applied the discovery algorithm to each of the databases mentioned above. We repeat the experiment with match thresholds ranging from 0.0 to 1.0 in increments of 0.1. Table 1 shows the description length (DL) of the original graph, the description length of the best substructure discovered by SUBDUE, and the value of compression. Compression here is defined as $\frac{\text{DL of compressed graph}}{\text{DL of original graph}}$. Figure 14, shows the actual discovered substructures for the first four datasets.

As can be seen from Table 1, SUBDUE was able to reduce the database to slightly larger than $\frac{1}{4}$ of its original size in the best case. The average compression value over all of these domains (treating the artificial graphs as one value) is 0.62. The results of this experiment demonstrate that the substructure discovered by SUBDUE can significantly reduce the amount of data needed to represent an input graph. We expect that compressing the graph using combinations of substructures and hierarchies of substructures will realize even greater compression in some databases.

| Database | $DL_{original}$ | $Threshold_{optimal}$ | $DL_{compressed}$ | Compression |
|---|---|---|---|---|
| Rubber | 371.78 | 0.1 | 95.20 | 0.26 |
| Cortisone | 355.03 | 0.3 | 173.25 | 0.49 |
| DNA | 2427.93 | 1.0 | 2211.87 | 0.91 |
| Pencils | 1592.33 | 1.0 | 769.18 | 0.48 |
| $CAD - M1$ | 4095.73 | 0.7 | 2148.8 | 0.52 |
| $CAD - S1SegDec$ | 1860.14 | 0.7 | 1149.29 | 0.62 |
| $CAD - S1DrvBlk$ | 12715.12 | 0.7 | 9070.21 | 0.71 |
| $CAD - BlankSub$ | 8606.69 | 0.7 | 6204.74 | 0.72 |
| $CAD - And2$ | 427.73 | 0.1 | 324.52 | 0.76 |
| Artificial (avg. over 96 graphs) | 1636.25 | 0.0...1.0 | 1164.02 | 0.71 |

Table 1: Graph compression results.

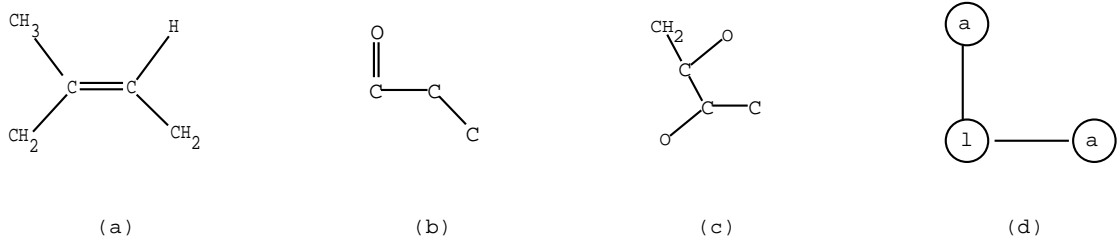

(a)  (b)  (c)  (d)

Figure 14: Best substructure for (a) rubber database, (b) cortisone database, (c) DNA database, and (d) image database.

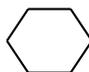

Figure 15: Benzene ring discovered by SUBDUE.

### 7.2.2 EXPERIMENT 2: RE-DISCOVERY OF KNOWN SUBSTRUCTURES USING BACKGROUND KNOWLEDGE

Another way of evaluating the discovery process is to evaluate the *interestingness* of the discovered substructures. The determination of this value will change from domain to domain. As a result, in this second set of experiments we test SUBDUE's ability to discover substructures that have already been labeled as important by experts in the domains under consideration.

In the chemical compound domain, chemists frequently describe compounds in terms of the building-block components that are heavily used. For example, in the rubber compound database shown in Figure 7, the compound is made up of a chain of structures that are labeled by chemists as *isoprene* units. SUBDUE's ability to re-discover this structure is exemplified in Figure 14a. This substructure, which was discovered using the MDL principle with no extra background knowledge, represents an isoprene unit.

Although SUBDUE was able to re-discover isoprene units without extra background knowledge, the substructure affording the most compression will not always be the most interesting or important substructure in the database. For example, in the cortisone database the *benzene ring* which consists of a ring of carbons is not discovered using only the MDL principle. However, the additional background rules can be used to increase the chance of finding interesting substructures in these domains. In the case of the cortisone compound, we know that the interesting structures exhibit a characteristic of closure. Therefore, we give a strong weight (8.0) to the *compactness* background rule and use a match threshold of 0.2 to allow for deviations in the benzene ring instances. In the resulting output, SUBDUE finds the benzene ring shown in Figure 15.

In the same way, we can use the background rules to find the pencil substructure in the image data. When the image in Figure 9 is viewed, the substructure of interest is the pencil in its various forms. However, the substructure that afforded the most compression does not make up an entire pencil. We know that the pencils have a high degree of closure and of coverage, so the weights for these rules are set to 1.0. With these weights, SUBDUE is able to find the pencil substructure shown in Figure 16 for all tested match thresholds between 0.0 and 1.0.

## 8. Hierarchical Concept Discovery

After a substructure is discovered, each instance of the substructure in the input graph can be replaced by a single vertex representing the entire substructure. The discovery procedure can then be repeated on the compressed data set, resulting in new interesting substructures. If the newly-discovered substructures are defined in terms of existing substructure concepts, the substructure definitions form a hierarchy of substructure concepts.

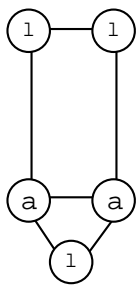

Figure 16: Pencil substructure discovered by SUBDUE.

Hierarchical concept discovery also adds the capability to improve SUBDUE's performance. When SUBDUE is applied to a large input graph, the complexity of the algorithm prevents consideration of larger substructures. Using hierarchical concept discovery, SUBDUE can first discover those smaller substructures which best compress the data. Applying the compression reduces the graph to a more manageable size, increasing the chance that SUBDUE will find the larger substructures on the subsequent passes through the database.

Once SUBDUE selects a substructure, all vertices that comprise the exact instances of the substructure are replaced in the graph by a single vertex representing the discovered substructure. Edges connecting vertices outside the instance to vertices inside the instance now connect to the new vertex. Edges internal to the instance are removed. The discovery process is then applied to the compressed data. If a hierarchical description of concepts is particularly desired, heavier weight can be given to substructures which utilize previously discovered substructures. The increased weight reflects increased attention to this substructure. Figure 17 illustrates the compressed rubber compound graph using the substructure shown in Figure 14a.

To demonstrate the ability of SUBDUE to find a hierarchy of substructures, we let the system make multiple passes through a database that represents a portion of a DNA molecule. Figure 8 shows a portion of two chains of a double helix, using three pairs of bases which are held together by hydrogen bonds. Figure 18 shows the substructures found by SUBDUE after each of three passes through the data. Note that, on the third pass, SUBDUE linked together the instances of the substructure in the second pass to find the chains of the double helix.

Although replacing portions of the input graph with the discovered substructures compresses the data and provides a basis for discovering hierarchical concepts in the data, the substructure replacement procedure becomes more complicated when concepts with inexact instances are discovered. When inexact instances of a discovered concept are replaced by a single vertex in the data, all distortions of the graph (differences between the instance graph and the substructure definition) must be attached as annotations to the vertex label.

Highest-valued substructure

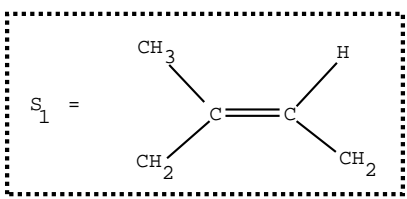

Compressed graph using discovered substructure

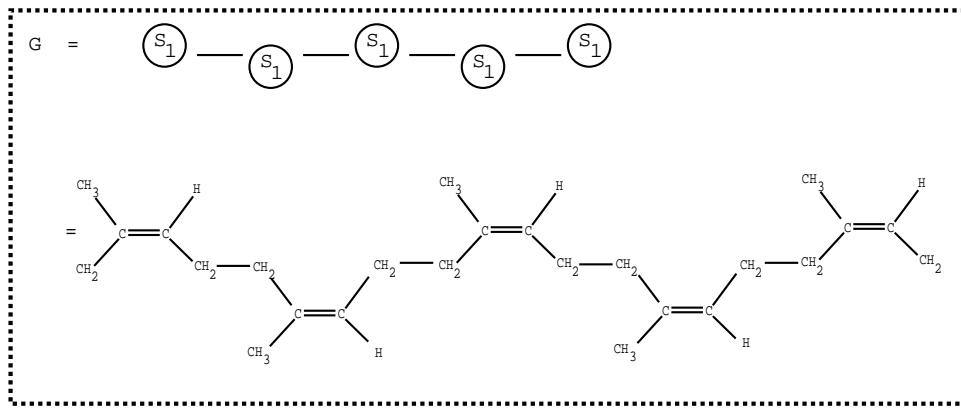

Figure 17: Compressed graph for rubber compound data.

**Highest-valued substructure
after First Pass**

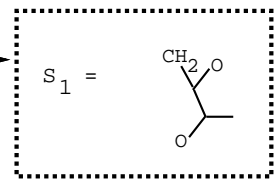

**Highest-valued substructure
after Second Pass**

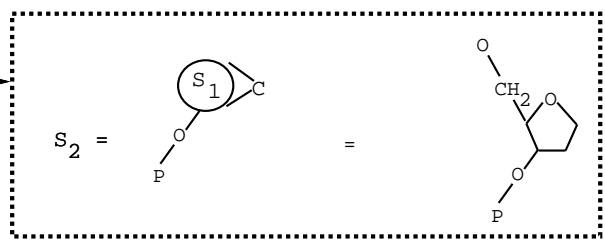

**Highest-valued substructure
after Third Pass**

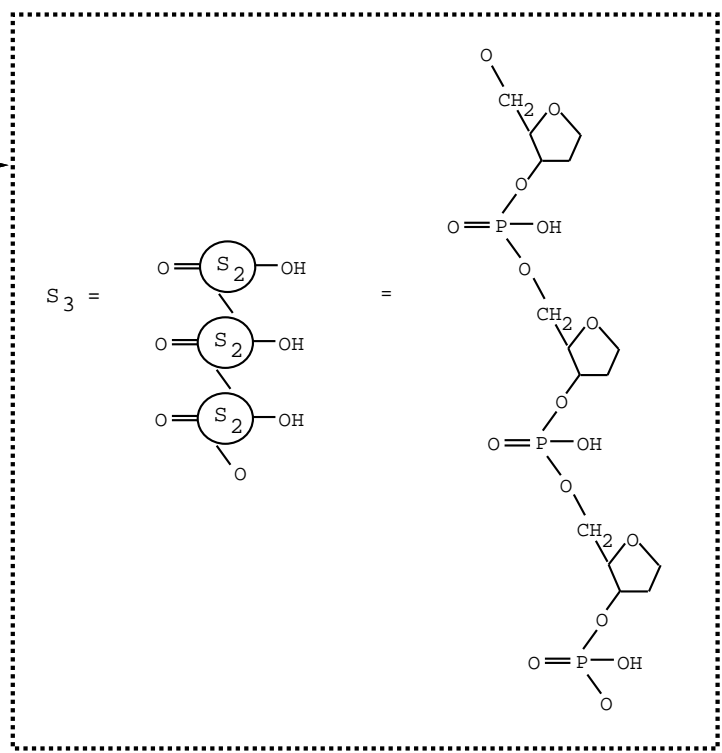

Figure 18: Hierarchical discovery in DNA data.

## 9. Conclusions

Extracting knowledge from structural databases requires the identification of repetitive substructures in the data. Substructure discovery identifies interesting and repetitive structure in structural data. The substructures represent concepts found in the data and a means of reducing the complexity of the representation by abstracting over instances of the substructure. We have shown how the minimum description length (MDL) principle can be used to perform substructure discovery in a variety of domains. The substructure discovery process can also be guided by background knowledge. The use of an inexact graph match allows deviation in the instances of a substructure. Once a substructure is discovered, instances of the substructure can be replaced by the concept definition, affording compression of the data description and providing a basis for discovering hierarchically-defined structures.

Future work will combine structural discovery with discovery of concepts using a linear-based representation such as AutoClass (Cheeseman, Kelly, Self, Stutz, Taylor, & Freeman, 1988). In particular, we will use Subdue to compress the data fed to AutoClass, and let Subdue evaluate the interesting structures in the classes generated by AutoClass. In addition, we will be developing a parallel implementation of the AutoClass / Subdue system that will enable application of substructure discovery to larger structural databases.

## Acknowledgements

This project is supported by NASA grant NAS5-32337. The authors would like to thank Mike Shay at National Semiconductor for providing the circuit data. We would also like to thank Surnjani Djoko and Tom Lai for their help with this project. Thanks also to the reviewers for their numerous insightful comments.

## References

Bunke, H., & Allermann, G. (1983). Inexact graph matching for structural pattern recognition. *Pattern Recognition Letters, 1*(4), 245–253.

Cheeseman, P., Kelly, J., Self, M., Stutz, J., Taylor, W., & Freeman, D. (1988). Autoclass: A bayesian classification system. In *Proceedings of the Fifth International Workshop on Machine Learning*, pp. 54–64.

Conklin, D., & Glasgow, J. (1992). Spatial analogy and subsumption. In *Proceedings of the Ninth International Machine Learning Workshop*, pp. 111–116.

Derthick, M. (1991). A minimal encoding approach to feature discovery. In *Proceedings of the Ninth National Conference on Artificial Intelligence*, pp. 565–571.

Fisher, D. H. (1987). Knowledge acquisition via incremental conceptual clustering. *Machine Learning, 2*(2), 139–172.

Fu, K. S. (1982). *Syntactic Pattern Recognition and Applications.* Prentice-Hall.

Holder, L. B., Cook, D. J., & Bunke, H. (1992). Fuzzy substructure discovery. In *Proceedings of the Ninth International Machine Learning Conference*, pp. 218–223.

Holder, L. B., & Cook, D. J. (1993). Discovery of inexact concepts from structural data. *IEEE Transactions on Knowledge and Data Engineering, 5*(6), 992–994.

Jeltsch, E., & Kreowski, H. J. (1991). Grammatical inference based on hyperedge replacement. In *Fourth International Workshop on Graph Grammars and Their Application to Computer Science*, pp. 461–474.

Leclerc, Y. G. (1989). Constructing simple stable descriptions for image partitioning. *International journal of Computer Vision, 3*(1), 73–102.

Levinson, R. (1984). A self-organizing retrieval system for graphs. In *Proceedings of the Second National Conference on Artificial Intelligence*, pp. 203–206.

Michalski, R. S., & Stepp, R. E. (1983). Learning from observation: Conceptual clustering. In Michalski, R. S., Carbonell, J. G., & Mitchell, T. M. (Eds.), *Machine Learning: An Artificial Intelligence Approach, Vol. I*, pp. 331–363. Tioga Publishing Company.

Miclet, L. (1986). *Structural Methods in Pattern Recognition*. Chapman and Hall.

Pednault, E. P. D. (1989). Some experiments in applying inductive inference principles to surfa ce reconstruction. In *Proceedings of the International Joint Conference on Artificial Intelligence*, pp. 1603–1609.

Pentland, A. (1989). Part segmentation for object recognition. *Neural Computation, 1*, 82–91.

Prather, R. (1976). *Discrete Mathemetical Structures for Computer Science*. Houghton Miffin Company.

Quinlan, J. R., & Rivest, R. L. (1989). Inferring decision trees using the minimum description length principle. *Information and Computation, 80*, 227–248.

Rao, R. B., & Lu, S. C. (1992). Learning engineering models with the minimum description length principle. In *Proceedings of the Tenth National Conference on Artificial Intelligence*, pp. 717–722.

Rissanen, J. (1989). *Stochastic Complexity in Statistical Inquiry*. World Scientific Publishing Company.

Schalkoff, R. J. (1992). *Pattern Recognition: Statistical, Structural and Neural Approaches*. John Wiley & Sons.

Segen, J. (1990). Graph clustering and model learning by data compression. In *Proceedings of the Seventh International Machine Learning Workshop*, pp. 93–101.

Thompson, K., & Langley, P. (1991). Concept formation in structured domains. In Fisher, D. H., & Pazzani, M. (Eds.), *Concept Formation: Knowledge and Experience in Unsupervised Learning*, chap. 5. Morgan Kaufmann Publishers, Inc.

Waltz, D. (1975). Understanding line drawings of scenes with shadows. In Winston, P. H. (Ed.), *The Psychology of Computer Vision*. McGraw-Hill.

Wertheimer, M. (1939). Laws of organization in perceptual forms. In Ellis, W. D. (Ed.), *A Sourcebook of Gestalt Psychology*, pp. 331–363. Harcourt, Brace and Company.

Winston, P. H. (1975). Learning structural descriptions from examples. In Winston, P. H. (Ed.), *The Psychology of Computer Vision*, pp. 157–210. McGraw-Hill.

Yoshida, K., Motoda, H., & Indurkhya, N. (1993). Unifying learning methods by colored digraphs. In *Proceedings of the Learning and Knowledge Acquisition Workshop at IJCAI-93*.

Zahn, C. T. (1971). Graph-theoretical methods for detecting and describing gestalt clusters. *IEEE Transactions on Computers, 20*(1), 68–86.